\documentclass[runningheads]{llncs}
\usepackage[T1]{fontenc}
\usepackage{graphicx}
\usepackage{cite}
\usepackage{amsmath}
\usepackage{amssymb}
\begin{document}
%

\newcommand{\corrauthor}{\textsuperscript{*}}

\title{CoSSeg-TTA: Contrast-Aware Semi-Supervised Segmentation with Domain Generalization and Test-Time Adaptation}
\titlerunning{Semi-Supervised Liver Segmentation}              
%
%
\author{Jincan Lou\inst{1} \and
Jingkun Chen\inst{3}\corrauthor \and
Haoquan Li\inst{1} \and
Hang Li \inst{1} \and
Wenjian Huang\inst{1} \and
Weihua Chen\inst{2} \and
Fan Wang\inst{2} \and
Jianguo Zhang\inst{1}\corrauthor}
\authorrunning{Jincan et al.}
%
\institute{Southern University of Science and Technology, Guangdong, China \and DAMO Academy, China \and
Department of Engineering Science, University of Oxford, Oxford, UK\\[2pt]
}
%

%
%
%

\maketitle              
%

\renewcommand{\thefootnote}{}
\footnotetext{\corrauthor Corresponding authors: Jingkun Chen \texttt{jingkun.chen@eng.ox.ac.uk}, Jianguo Zhang \texttt{zhangjg@sustech.edu.cn}}
\renewcommand{\thefootnote}{\arabic{footnote}}

\begin{abstract}
Accurate liver segmentation from contrast-enhanced MRI is essential for diagnosis, treatment planning, and disease monitoring. However, it remains challenging due to limited annotated data, heterogeneous enhancement protocols, and significant domain shifts across scanners and institutions. Traditional image-to-image translation frameworks have made great progress in domain generalization, but their application is not straightforward. For example, Pix2Pix requires image registration, and cycle-GAN cannot be integrated seamlessly into segmentation pipelines. Meanwhile, these methods are originally used to deal with cross-modality scenarios, and often introduce structural distortions and suffer from unstable training, which may pose drawbacks in our single-modality scenario. To address these challenges, we propose CoSSeg-TTA, a compact segmentation framework for the GED4 (Gd-EOB-DTPA enhanced hepatobiliary phase MRI) modality built upon nnU-Netv2 and enhanced with a semi-supervised mean teacher scheme to exploit large amounts of unlabeled volumes. A domain adaptation module, incorporating a randomized histogram-based style appearance transfer function and a trainable contrast-aware network, enriches domain diversity and mitigates cross-center variability. Furthermore, a continual test-time adaptation strategy is employed to improve robustness during inference. Extensive experiments demonstrate that our framework consistently outperforms the nnU-Netv2 baseline, achieving superior Dice score and Hausdorff Distance while exhibiting strong generalization to unseen domains under low-annotation conditions.

\end{abstract}

\section{Introduction}
Liver fibrosis, a common consequence of chronic liver injuries, can progress to cirrhosis, liver failure, or hepatocellular carcinoma, posing serious global health concerns. Accurate liver segmentation is essential in clinical workflows, including diagnosis, treatment planning, and surgical navigation. However, acquiring high-quality annotations is labor-intensive, requires expert knowledge, and is prone to inter-observer variability, which often leads to annotation scarcity.~\cite{marinov2024deep, rajchl2017employing}. 

Beyond annotation scarcity, two major technical challenges remain. First, clinical liver assessment always uses contrast-enhanced MRI, such as Gd-EOB-DTPA-enhanced imaging, but it suffers from artifacts and uneven enhancement~\cite{li2022domain, sendra2022domain, chen2020deep}. Second, modern clinical datasets are often collected from multiple centers and scanners, leading to severe domain shifts due to variations in vendors, acquisition protocols, and noise characteristics~\cite{vorontsov2022towards, nichyporuk2022rethinking, sendra2022domain}. 

Recent advances such as GAN-based domain transfer and image-to-image translation exhibit strong cross-domain generative capability. However, these approaches have critical limitations. Methods like Pix2Pix~\cite{isola2017image} always require image registration, and those like cycle-GAN~\cite{zhu2017unpaired} cannot be integrated seamlessly into our framework. Meanwhile, a large body of work uses GANs for cross-modality tasks, rather than for small scanner- or protocol-induced shifts within modality~\cite{zhang2022multi} , and will introduce artifacts and blur tissue edges. Their behavior is not ideal for our setting, which is domain shift within the same modality.

To address these issues, we propose a semi-supervised segmentation framework that integrates domain generalization and test-time adaptation. Our approach builds on nnU-Netv2~\cite{isensee2021nnu} under mean teacher paradigm with pseudo-labels learning to solve annotation scarcity. A pre-processing module combines domain transfer method with contrast-aware network, while continual test-time adaptation (CoTTA)~\cite{wang2022continual} improves robustness to unseen domains. Finally post-processing is applied to trim the prediction mask.

Our contributions are threefold:  
1) We design a semi-supervised framework based on nnU-Netv2 under mean teacher strategy with pseudo-labels learning, effectively leveraging unlabeled dataset.
2) We introduce a domain adaptation pipeline combining appearance translation via random histogram matching and a contrast-aware network, integrated with test-time adaptation to improve cross-domain generalization.  
3) Our approach is validated on a real-world multi-center dataset, demonstrating superior robustness and accuracy compared to existing methods in low-label and cross-domain scenarios.

\section{Method}

\begin{figure}[!]
    \centering
    \includegraphics[width=0.9\textwidth]{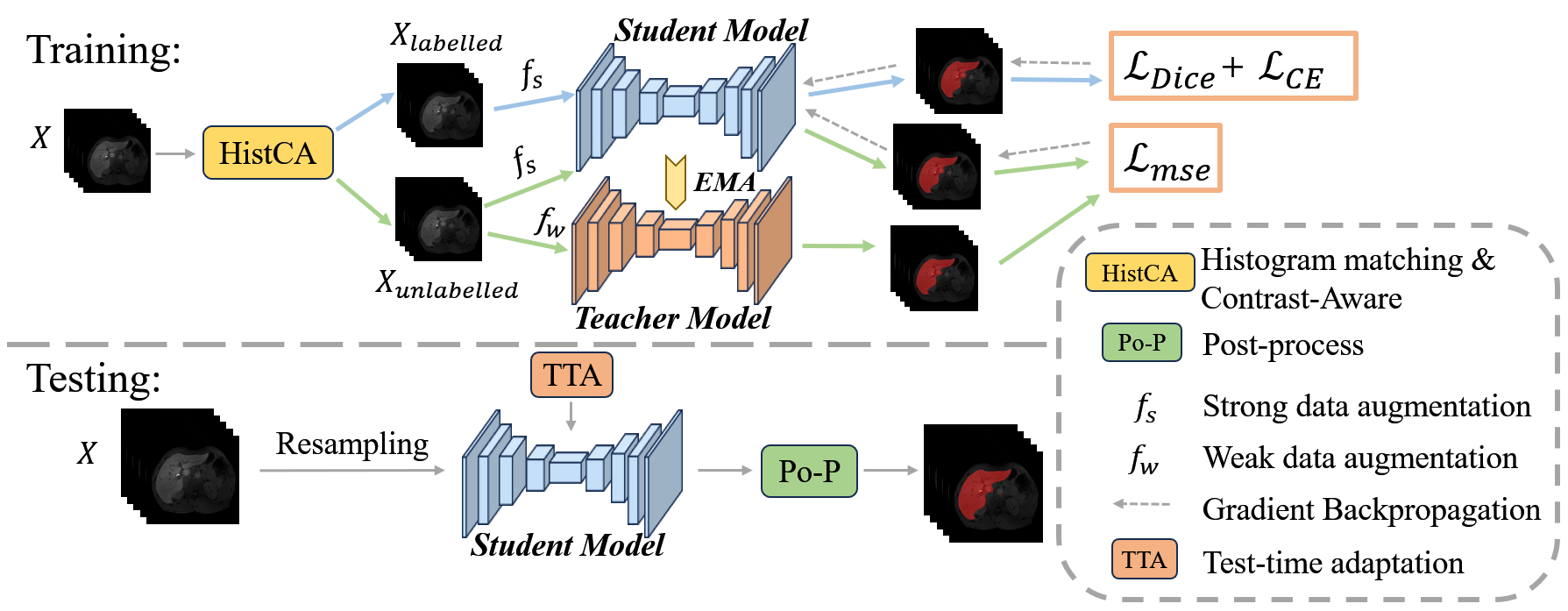}
    \caption{An overview of the proposed semi-supervised liver segmentation framework.}
    \label{fig:framework}
\end{figure}

Fig.~\ref{fig:framework} illustrates the overall pipeline: during training, the model applies pre-processing techniques, including domain adaptation, contrast enhancement, and data augmentation, and is trained under the framework of mean teacher with pseudo-labels learning; during inference, continual test-time adaptation and post-processing further improve robustness.

\subsection{Semi-Supervised Learning with mean teacher}
Since manual annotations are scarce and costly to obtain in clinical practice \cite{chen2025addressing, zhang2025helpnet, chen2025gaze}, we adopt a semi-supervised learning framework based on the mean teacher paradigm~\cite{tarvainen2017mean, chen2023semi}. This approach enables the model to effectively leverage large amounts of unlabeled data alongside a limited set of labeled examples \cite{chen2024dynamic}.

The framework consists of two models: a student network $h_\theta$ with parameters $\theta$, and a teacher network $h_\phi$ with parameters $\phi$. The teacher is updated as the exponential moving average (EMA) of the student parameters:

\begin{equation}
\phi \leftarrow \alpha \phi + (1 - \alpha)\theta
\end{equation}

where $\alpha \in [0,1)$ is a smoothing coefficient that controls the update rate. This strategy ensures that the teacher evolves more smoothly, providing stable targets for consistency training.

Given a labeled training dataset $\mathcal{D}_l^s$ and an unlabeled training dataset $\mathcal{D}_u^s$, the overall training objective is formulated as:
\begin{align}
\mathcal{L} = \mathcal{L}_{\text{sup}}(\mathcal{D}_l^s) + \lambda_{\text{mse}} \cdot \mathcal{L}_{\text{mse}}(\mathcal{D}_u^s) \\
\mathcal{L}_{\text{sup}} = \mathcal{L}_{\text{Dice}} + \mathcal{L}_{\text{ce}}(\mathcal{D}_l^s)
\end{align}

where $\mathcal{L}_{\text{sup}}$ is the supervised segmentation loss to learn from labeled dataset. $\mathcal{L}_{\text{mse}}$, which is mean square error loss, is used as consistency loss that enforces agreement between student and teacher predictions on unlabeled data under different perturbations.

In practice, we apply strong data augmentation (e.g., histogram matching, geometric transformations) to generate two perturbed views of each unlabeled sample, encouraging the student to produce consistent outputs with the teacher. This design effectively exploits unlabeled data, improves feature representation learning, and mitigates overfitting in low-annotation settings.

\subsection{Domain Adaptation via Appearance translation}
In multi-domain medical imaging, the marginal input distribution $P(x)$ often varies significantly across centers and scanners, a phenomenon known as covariate shift. If unaddressed, this distribution discrepancy causes poor cross-domain generalization. 

\begin{figure}[!]
    \centering
    \includegraphics[width=0.9\textwidth]{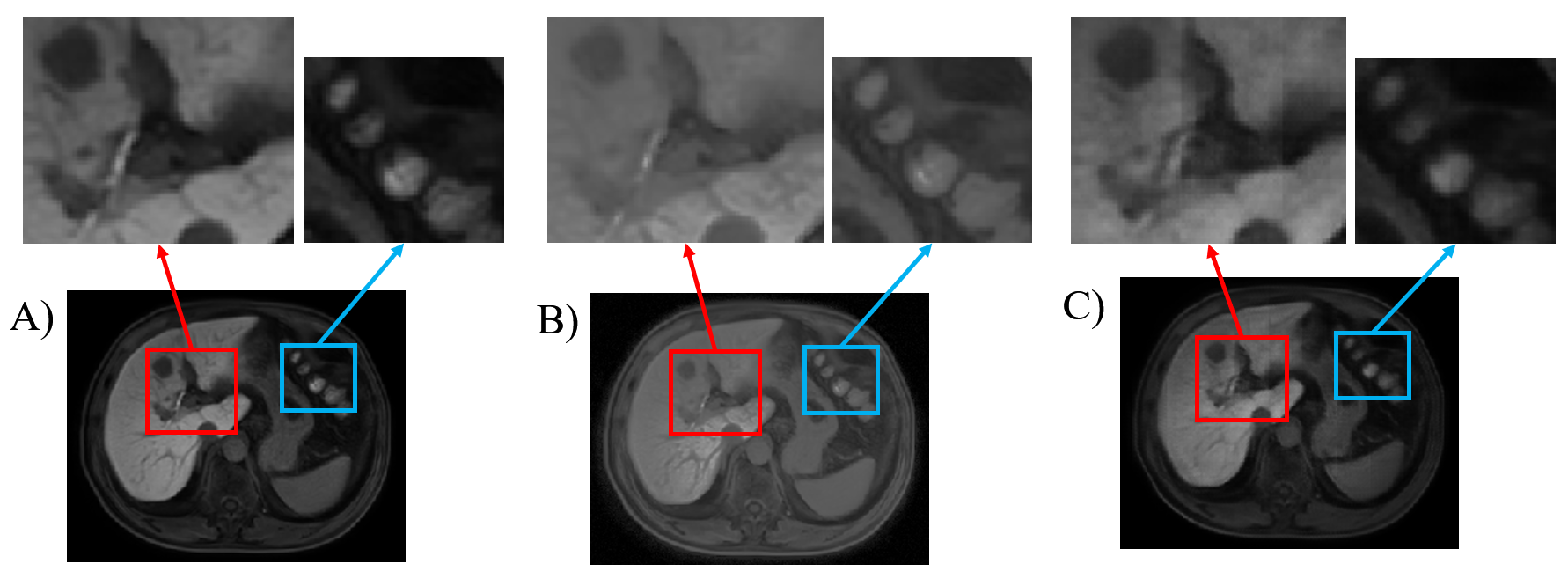}
    \caption{A) Original GED4 modality image; B) Style transferred with random matching histogram; C) Style transferred with GAN. The red triangle areas show the comparison of artifacts, while the blue triangle areas show the comparison of blurred pattern}
    \label{fig:hist}
\end{figure}

Instead of GAN based methods, we adapt a lightweight appearance translation strategy based on histogram matching, which can be directly embedded into data pre-processing and training. The method of random histogram matching efficiently deal with domain generalization, and avoid the generation of artifacts and blurred pattern (see Fig.~\ref{fig:hist}).

We also provide the principle of Histogram Matching, and prove its efficiency in domain generalization (see Appendix).

\subsection{Contrast-aware Module}
To address the inconsistent visibility of liver structures across contrast phases, we design a learning-based contrast-aware module that exploits the contrast relationship between T1-weighted modality (T1WI) and GED4 modalities.

In contrast-enhanced medical imaging, different acquisition phases often exhibit significant differences in soft tissue visibility and organ boundary clarity. Specifically, in our dataset, T1WI corresponds to the pre-contrast phase, while the GED4 modality represents the final post-contrast enhanced phase, where liver lesions and vascular structures are more conspicuously visualized due to contrast agent accumulation. Motivated by this inherent contrast evolution, we propose a learning-based module that captures the mapping between T1WI and GED4 modalities to enhance the contrast representation in GED4 images.

We hypothesize that the contrast relationship between the T1WI and GED4 phases can be exploited using deep learning to generate contrast-enhanced representations that emphasize relevant anatomical structures. To this end, we first perform rigid registration and 3D alignment between T1WI and GED4 volumes using the FSL-Flirt toolkit~\cite{jenkinson2002improved}. Subsequently, all volumes are resampled to a unified spatial resolution of $1.0 \times 1.0 \times 2.5$ mm\textsuperscript{3} to ensure spatial consistency and facilitate voxel-wise correspondence learning.

A 3D U-Net~\cite{cciccek20163d} is applied to learn the forward mapping function $G: X_{\text{T1}}\rightarrow \hat{X}_{\text{GED4}}$. The objective is to reconstruct a synthetic contrast-enhanced image $\hat{X}_{\text{GED4}}$ that closely approximates the  corresponding real $X_{\text{GED4}}$. 

The training loss for this module combines two components:

\begin{equation}
\mathcal{L}_{\text{con-enh}} = \mathcal{L}_{\text{MSE}} + \lambda_{\text{SSIM}} \cdot \left(1 - \text{SSIM}( \hat{X}_{\text{GED4}}, X_{\text{GED4}} )\right)
\end{equation}

where $\mathcal{L}_{\text{MSE}}$ is the mean squared error between the synthetic and real GED4 images, and $\text{SSIM}$~\cite{wang2004image} is the structural similarity index that encourages perceptual and textural consistency. This combination ensures that the reconstructed images are not only pixel-wise accurate but also preserve high-frequency anatomical structures relevant to contrast uptake (see Appendix).


\subsection{Pseudo-Labels Learning}
In addition to the semi-supervised learning approach, we further enhance the model's ability to generalize by leveraging pseudo-labels. Pseudo-labeling is an effective strategy to exploit unlabeled data by using the model’s own predictions as labels. In our framework, after training the mean teacher model on the training set, we apply the trained student model to segment the new unlabeled dataset, generating pseudo-labels for each new sample.

These pseudo-labels are generated under the assumption that the model has already learned meaningful features from the training set data and are then used to augment the original training set. We combine the training set $\mathcal{D}^s$ with the pseudo-labeled new set $\mathcal{D}^{new}_{\text{pseudo}}$ to fine-tune the model. The fine-tuning process incorporates both the true labels and pseudo-labels, refining the model's parameters to improve its generalization to unseen data. This final fine-tuning step strengthens the model’s robustness by exposing it to a larger variety of data, enhancing segmentation accuracy and generalization.

Formally, the new fine-tuning objective becomes:
\begin{equation}
\mathcal{L}_{\text{finetune}} = \mathcal{L}_{\text{sup}}(\mathcal{D}_l^s \cup \mathcal{D}^{new}_{\text{pseudo}}) + \lambda_{\text{mse}} \cdot \mathcal{L}_{\text{mse}}(\mathcal{D}_u^s)
\end{equation}

\subsection{Test-Time Adaptation Module and post-processing}
Even with domain adaptation during training, unseen test domains may still introduce significant distribution shifts due to scanner-dependent noise patterns or different acquisition protocols. 
To further enhance robustness at inference time, we integrate a test-time adaptation (TTA) module with CoTTA~\cite{wang2022continual} into our framework. 
Unlike static domain transfer methods, CoTTA adapts the model online by updating its parameters as new test samples are encountered, without requiring access to the source training data.

Specifically, given a pretrained student model $h_{\theta}$, CoTTA maintains an exponential moving average (EMA) teacher model $h_{\phi}$ to provide stable pseudo-labels. 
For each incoming test volume $x_t$, the model prediction $\hat{y}_t = h_{\theta}(x_t)$ is refined by consistency regularization with the teacher output $h_{\phi}(x_t)$. 
At the same time, stochastic restoration is applied to a subset of network parameters, preventing error accumulation and catastrophic forgetting during continual adaptation. 

The overall objective at test time can be expressed as:
\begin{equation}
\mathcal{L}_{\text{TTA}} = \| h_{\theta}(x_t) - h_{\phi}(x_t) \|^2,
\end{equation}
Allowing $\theta$ to adapt progressively to the target distribution. 

However, the predicted mask may still surrounding with noise points or areas. To further address this issue, we apply 3D morphological post-processing to further trim the segmentation mask.

\section{Experiment}
\subsection{Dataset and metrics}


The dataset utilizes the CARE-Liver track of CARE 2025, collected from multiple clinical centers using different MRI scanners~\cite{liu2025merit, gao2023reliable, wu2022meru}. It includes a training set with 330 unlabeled and 30 labeled cases, a validation set with 60 unlabeled cases, and an unseen test set with 60 labeled cases. Each case includes 7 imaging modalities, while in our framework, we only require T1WI and GED4 modalities among them.

During the training phase, the whole training set is used during the pre-train stage for semi-supervised learning, and whole the validation set is used during the finetune stage for pseudo-labels learning. For each experiment, we use five fold cross validation to split these 30 labeled cases for evaluating experiment. 

To quantitatively evaluate the performance of liver segmentation, we adopt two widely used metrics in medical image analysis: Dice score (DICE) and standard Hausdorff Distance (HD). The HD is computed in the physical space with millimeters (mm) as the unit. We report the average performance of five fold cross validation on each method.

\subsection{Setup}

All experiments were run on a single RTX 3090 GPU. We use the same epochs of $150$ and learning rate of $0.01$ with the same scheduler policy for each training. During the ablation experiments, for mean teacher frame, $\lambda_{\text{mse}}$ is dynamically increased from $0.0$ to $1.0$ with a ramp-up step of $40$ epochs. During the training of contrast-aware module, we set $\lambda_{\text{SSIM}}$ as $0.8$.

For the training stage, we first use aligned T1WI–GED4 pairs to train the contrast-aware module. Once trained, this module is applied to the input GED4 data to generate enhanced GED4 volumes. The original and enhanced GED4 images are subsequently concatenated as a two-channel input to the nnU-Netv2 model. Before feeding into the network, random histogram matching and standard data augmentation strategies are applied. The segmentation model is trained under the mean teacher framework with pseudo-labels learning to leverage both labeled and unlabeled data.

For the inference stage, the trained student model produces segmentation predictions, which are further refined by the post-processing module to remove small false positives and fragmented regions. In addition, when test samples originate from previously unseen domains, the CoTTA module is activated to dynamically adapt the model to the target distribution during deployment.

\subsection{Segmentation results analysis}
\subsubsection{Baseline Comparison}

\begin{table}[ht!]
\centering
\caption{Comparison between different baselines.}\label{baseline}
\begin{tabular}{|c|c|c|c|c|}
\hline 
Metrics &  U-Net~\cite{cciccek20163d} & VNet~\cite{milletari2016v}  & nnU-Netv2~\cite{isensee2021nnu} & Swin UNETR~\cite{hatamizadeh2021swin} \\
\hline 
DICE &  95.01 & 94.83 & \textbf{95.78} & 93.63 \\
HD &  75.06 & 66.69 & 69.35 & \textbf{42.64} \\
\hline
\end{tabular}
\end{table}

We first use labeled training set to evaluate the segmentation performance of different classic baseline models and select the best baseline model without TTA and post-processing strategies. As shown in Table~\ref{baseline}, the nnU-Netv2~\cite{isensee2021nnu} model achieved the highest DICE score of 95.78, surpassing other baselines such as U-Net~\cite{cciccek20163d}, VNet~\cite{milletari2016v}, and Swin UNETR~\cite{hatamizadeh2021swin}. Note that the maximum HD is highly sensitive to single outliers and does not fully capture overall boundary quality. Thus, although nnU-Netv2 had a higher HD score (69.35), the primary focus of our study was on improving DICE, where nnU-Netv2 demonstrated the most promising results.

\begin{table}[ht!]
\centering
\caption{Ablation analysis of methods over cycle-GAN, random histogram matching, mean teacher framework, pseudo-labels learning, contrast-aware module, continual test-time adaptation and post-processing. The baseline model is nnU-Netv2. Mean DICE and HD are used to evaluate each method. Five fold cross validation is used to get the mean DICE and HD. A$\sim$G are local experiments result, while the last two rows are the challenge test phase results with an unseen domain dataset. ID* refers to the source domain test dataset, and OOD* refers to an unseen domain test dataset.}
\label{ablation}
\begin{tabular}{|cccccccc|c|c|}
\hline
Exp & GAN & HIST & Mean-T &  pseu-learn & Con-Aware & CoTTA & Post-P & DICE ${\uparrow}$ & HD ${\downarrow}$ \\ \hline
A &  &  &  &  &  &  &  & 95.78 & 69.35 \\ \hline
B & $\checkmark$ &  &  &  &  &  &  & 95.94 & 55.67 \\ \hline
C &  & $\checkmark$ &  &  &  &  &  & 96.30 & 70.45 \\ \hline
D &  & $\checkmark$ & $\checkmark$ &  &  &  &  & 96.46 & 60.88 \\ \hline
E &  & $\checkmark$ & $\checkmark$ & $\checkmark$ &  &  &  & 96.52 & 64.29 \\ \hline
F &  & $\checkmark$ & $\checkmark$ & $\checkmark$ &  $\checkmark$ &   &  & 96.67 & 55.24 \\ \hline
G &  & $\checkmark$ & $\checkmark$ & $\checkmark$ &  $\checkmark$ & $\checkmark$ &  & 96.85 & 64.64 \\ \hline
H &  & $\checkmark$ & $\checkmark$ & $\checkmark$ &  $\checkmark$ & $\checkmark$ & $\checkmark$ &\textbf{96.88} & \textbf{49.22} \\ \hline  \hline
ID* &  &  &  &  &  &  &  &  97.03 & 21.4 \\ \hline
OOD* &  &  &  &  &  &  &  &  97.61 & 15.53 \\ \hline
\end{tabular}
\end{table}

\subsubsection{Ablation analysis}

In the ablation study, various methods were incorporated into the nnU-Netv2 to assess their impact on model performance. The techniques evaluated include cycle-GAN, random histogram matching, mean teacher pseudo-labels learning, and the contrast-aware module. Table~\ref{ablation} summarizes the results of the ablation study. Each method was tested in combination, and the mean DICE and HD scores were used to quantify the performance.

From the results, we observe that the inclusion of both GAN and random histogram matching lead to an improvement compared to the baseline model. While in comparison, random histogram matching shows a better performance in DICE. Their different effects could have been foreseen in Fig.~\ref{fig:hist}, where cycle-GAN has a poor generalization ability and image quality.

\begin{figure}[!]
    \centering
    \includegraphics[width=0.8\textwidth]{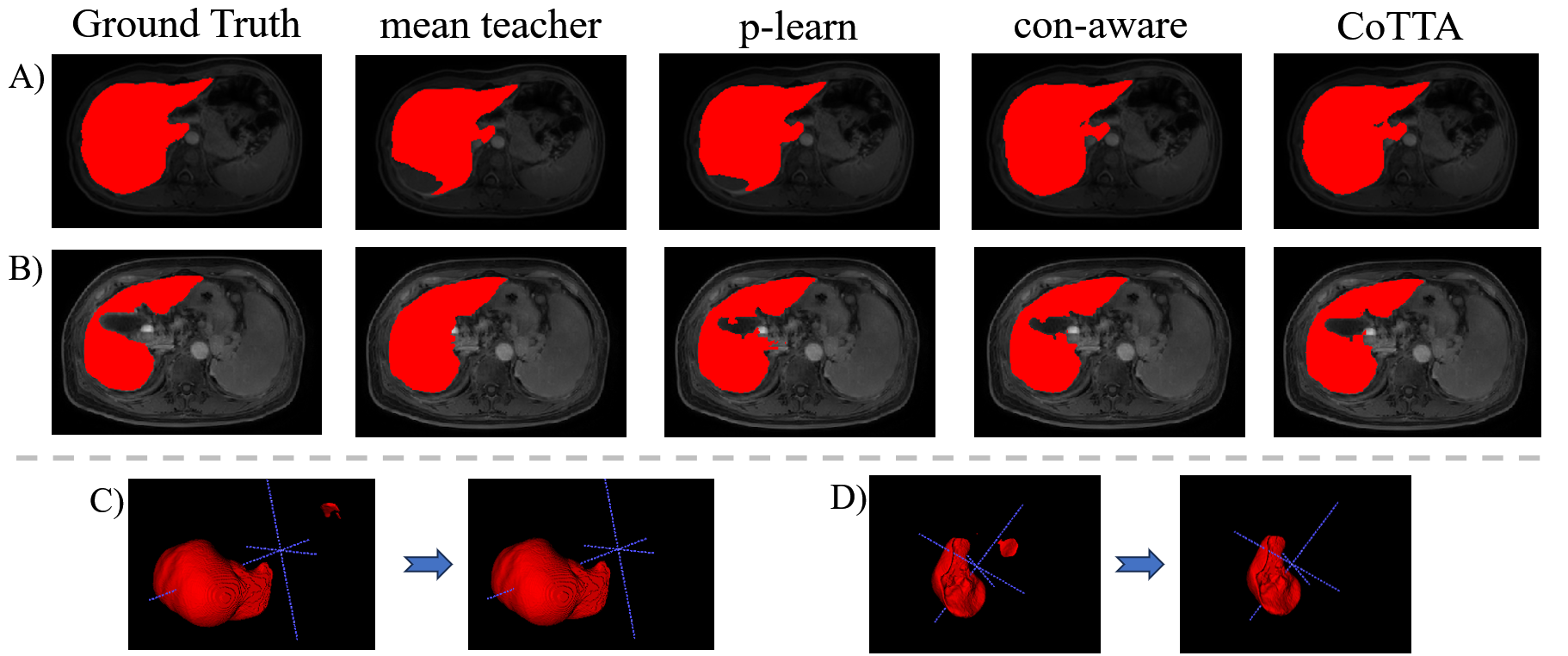}
    \caption{Visualization of typical segmentation results comparing different methods on the CARE-Liver track dataset of CARE 2025. In A) and B), from the $2nd$ to the $5th$, we add mean teacher, pseudo-labels learning, contrast-aware module and CoTTA in sequence. C) and D) show the optimization of segmentation results by post-processing module.}
    \label{fig:seg}
\end{figure}

During the training phase, we add the methods of mean teacher framework, pseudo-labels learning and contrast-aware module. The integration of mean teacher, pseudo-labels learning, and contrast-aware module resulted in a much better performance, which helps the model learn detailed knowledge about the liver region. CoTTA further improve the segmentation precision, and the model achieves the best performance with post-processing at DICE of 96.88 and HD of 49.22 in local experiments. Finally, the model achieves excellent results (ID*: 97.03 DICE, 21.4 HD; OOD*: 97.61 DICE, 15.53 HD) on the CARE 2025 challenge test phase. The visualization of some typical segmentation results by applying different methods is shown in Fig.~\ref{fig:seg}.

\section{Conclusion}

We proposed CoSSeg-TTA, a semi-supervised framework for liver segmentation from contrast-enhanced MRI, addressing annotation scarcity and domain shifts. Built on nnU-Netv2 under the mean teacher framework with pseudo-labels learning, it leverages unlabeled data, while a domain adaptation module with histogram-based style transfer and a contrast-aware network enhances segmentation robustness. CoTTA and post-processing further improve performance, achieving excellent results on the CARE 2025 challenge test phase. CoSSeg-TTA offers a robust solution for clinical liver segmentation tasks.

However, the contrast-aware module of our framework relies on paired T1WI and GED4 modalities, which may not always be available. Extreme domain shifts and noisy pseudo-labels can affect performance, and computational demands may limit real-time use. Future work includes extending to other modalities, improving the contrast-aware module with generative models, incorporating uncertainty estimation, and validating on larger datasets. Federated learning could address multi-center privacy concerns, enhancing clinical applicability.

\begin{credits}
\subsubsection{\ackname} This work is supported in part by National Natural Science Foundation of China (Grant No. 62276121), the TianYuan funds for Mathematics of the National Science Foundation of China (Grant No. 12326604).

\end{credits}

%
%
%
\bibliographystyle{splncs04}
\bibliography{ref}
%





\end{document}